\documentclass[conference,a4paper]{IEEEtran}
\usepackage{multirow}
\usepackage{graphicx}
\usepackage{amsmath}
\usepackage[psamsfonts]{amssymb}
\usepackage{amsxtra}
\usepackage{threeparttable}
\usepackage{tabularx}
\usepackage[justification=centering]{caption}
\usepackage{geometry}
\usepackage{fancyhdr}

\begin{document}

\title{Punctuation Restoration for Singaporean Spoken Languages: English, Malay, and Mandarin}

\author{
  Abhinav Rao\\BITS Pilani, Hyderabad Campus\\f20180172@hyderabad.bits-pilani.ac.in
  \and Ho Thi-Nga\\Nanyang Technological University\\ngaht@ntu.edu.sg
  \and Chng Eng-Siong\\Nanyang Technological University\\aseschng@ntu.edu.sg
}
\maketitle

\pagestyle{fancy}

\begin{abstract}
This paper presents the work of restoring punctuation for ASR transcripts generated by multilingual ASR systems. The focus languages are English, Mandarin, and Malay which are three of the most popular languages in Singapore. To the best of our knowledge, this is the first system that can tackle punctuation restoration for these three languages simultaneously.
Traditional approaches usually treat the task as a sequential labeling task, however, this work adopts a slot-filling approach that predicts the presence and type of punctuation marks at each word boundary. The approach is similar to the Masked-Language Model approach employed during the pre-training stages of BERT, but instead of predicting the masked word, our model predicts masked punctuation.
Additionally, we find that using  Jieba\footnote{https://github.com/fxsjy/jieba} instead of only using the built-in SentencePiece tokenizer of XLM-R can significantly improve the performance of punctuating Mandarin transcripts.

Experimental results on English and Mandarin IWSLT2022 datasets and Malay News show that the proposed approach achieved state-of-the-art results for Mandarin with 73.8\% F1-score while maintaining a reasonable F1-score for English and Malay, i.e. 74.7\% and 78\% respectively.
Our source code that allows reproducing the results and building a simple web-based application for demonstration purposes is available on Github\footnote{https://github.com/AetherPrior/Multilingual-Sentence-Boundary-detection}.
\end{abstract}

\section{Introduction}
Automatic punctuation restoration is a problem of restoring punctuation symbols on unpunctuated text. One of its common inputs is automatic transcripts generated by Automatic Speech Recognition (ASR) systems. This is because most current ASR systems still yield transcripts with no punctuation, which poses a challenge for human reading \cite{jones-et-al-punctutation} and also downstream text processing applications. Performance of text processing applications such as machine translation \cite{IWSLT2011/MT}, and sentence dependency parsing \cite{spitkovsky-etal-2011-punctuation}, has been shown to decrease when processed on unpunctuated text. Thus, restoring punctuation is crucial for not only improving the human readability of ASR-generated transcripts but also opening up their usability in many text processing applications.

There are two broad groups of input features that are used in the research on the task of punctuation restoration for ASR transcripts. One is prosody-based features that are extracted from speech output. For example, \cite{Annotate} uses pause duration and pitch with a Multi-Layer Perceptron (MLP) to predict punctuation marks after each word, while \cite{Batista2007RecoveringPM} supplements prosodic features with POS-tags for every word in the input sequence. \cite{tilk2015lstm} supplement their texts with pause durations, after aligning their audio data with text, and use an LSTM to punctuate texts. \cite{9747131} use an acoustic feature extractor, supplemented by BERT for lexical feature extraction. However, there are disadvantages of using prosody-based features in practical use, owing to the variations in tone per person and the non-availability of speech audio for different transcripts. The other is lexical features which are extracted purely from text sources. For example, \cite{article_HMM} uses a Hidden-Markov-Model approach, \cite{HMMInproceedings} uses the superior Conditional Random Field for punctuation restoration, and \cite{DBLP:conf/lrec/CheWYM16} uses word-vectors for embeddings. Punctuation restorers using only lexical features face a limitation of not utilizing audio information but are generally more straightforward and simple to implement.

Over the past few years, there has been increasing research in multilingual language models (MLM), with the advent of state-of-the-art models that can produce multilingual embeddings and can be fine-tuned for downstream tasks. Multilingual-BERT, a variation of BERT \cite{devlin2019bert} developed by Google, and XLM-RoBERTa \cite{conneau2020unsupervised} developed by Facebook are two such models that capture the vocabularies of over a hundred languages. These large transformer models are shipped with pre-trained weights and only require fine-tuning for other tasks such as Question-Answering, Named-Entity Recognition, and Natural Language Inference (NLI) with minimal modifications to the architecture.

In this work, we treat the punctuation prediction task as a slot-filling task, where, similar to the MLM objective, we consider the masking of punctuation marks as a fine-tuning step to predict punctuation. To be capable of performing the task in a multilingual setting, we use an XLM-RoBERTa (XLM-R) base model as the backbone pre-trained layer, and a classifier that outputs the punctuation marks to be filled in each masked position. XLM-R has shown to be performed well on Cross-Lingual tasks such as Cross-Lingual Natural Language Inference \cite{DBLP:XNLI}, Question-Answering \cite{DBLP:XQuAD}, and Information Retrieval \cite{tiedemann2020tatoeba}. Additionally, we modify the tokenization process and improve the model performance for the Mandarin test set. We evaluate this architecture for three languages, i.e. English, Mandarin, and Malay, that are popular in Singapore. \\

The rest of this paper is presented as follows. Section \ref{background} discusses the related work. Our proposed approach is described in Section \ref{proposed-model} with details about our model architecture and our analysis of the selected datasets. Experiments and results are presented in Section \ref{experiment}. We interpret our findings in Section \ref{discussion}. Section \ref{conclusion} concludes our work. 

\section{Background}
\label{background}
Punctuation restoration using Deep Learning has had its fair share of development over the past few years. \textit{Tilk et al.} \cite{tilk2015lstm}, \cite{tilk2016bidirectional} used an LSTM-based approach for restoring punctuation using sequence tagging, with \cite{salloum-etal-2017-deep} using a similar architecture for applying punctuation restoration to medical reports. Upon the advent of transformers \cite{DBLP:journals/corr/VaswaniSPUJGKP17}, \textit{Makhija et al.} \cite{bertpunct} use a BERT \cite{devlin2019bert} model as an embedding generator and an LSTM+CRF for punctuation restoration. \textit{Yi et al.} \cite{yi2020adversarial} supplement the punctuation prediction task with POS tagging for adversarial training and \cite{DiscriminQian} employ a teacher-student model for punctuation prediction, similar to knowledge distillation. \textit{Zhang et al.} \cite{ZheZhang} focus on Mandarin, capturing Mandarin word-boundary information using a CNN, while labelling the punctuation marks using BERT-Chinese combined with a sequence model. While all of the above models treat punctuation restoration as a sequence modeling task, SAPR \cite{SAPR} treats the problem as a machine translation task, and uses a transformer-based model for generating embeddings for the model.

Work has also been developed towards multilingual approaches: \textit{Alam et al.} \cite{TransformerLow} train multiple models for English and a lower-resource Indian language, Bangla, and \textit{A. Nagy. et al}. \cite{DBLP:journals/corr/ANagy} train for English and Hungarian, using English-BERT and Hungarian-BERT respectively. \textit{Chordia} \cite{chordia-2021-punktuator} separately trains and compares multilingual models for high resource languages such as English, Hindi, and low resource Indian languages. \textit{Guerreiro et al.} \cite{GUERREIRO2021115740} train and compare a multilingual and a monolingual punctuator using BERT models for European languages. \textit{Li et al.} \cite{li20m_interspeech} use a single LSTM model for forty-three different languages, with a modification to the tokenizer and vocabulary by using a byte-pair encoding (BPE) scheme to capture shared instances of vocabulary. 

\section{Proposed Model}
\label{proposed-model}
\begin{figure}[t]
	\centering
	\includegraphics[width=\linewidth]{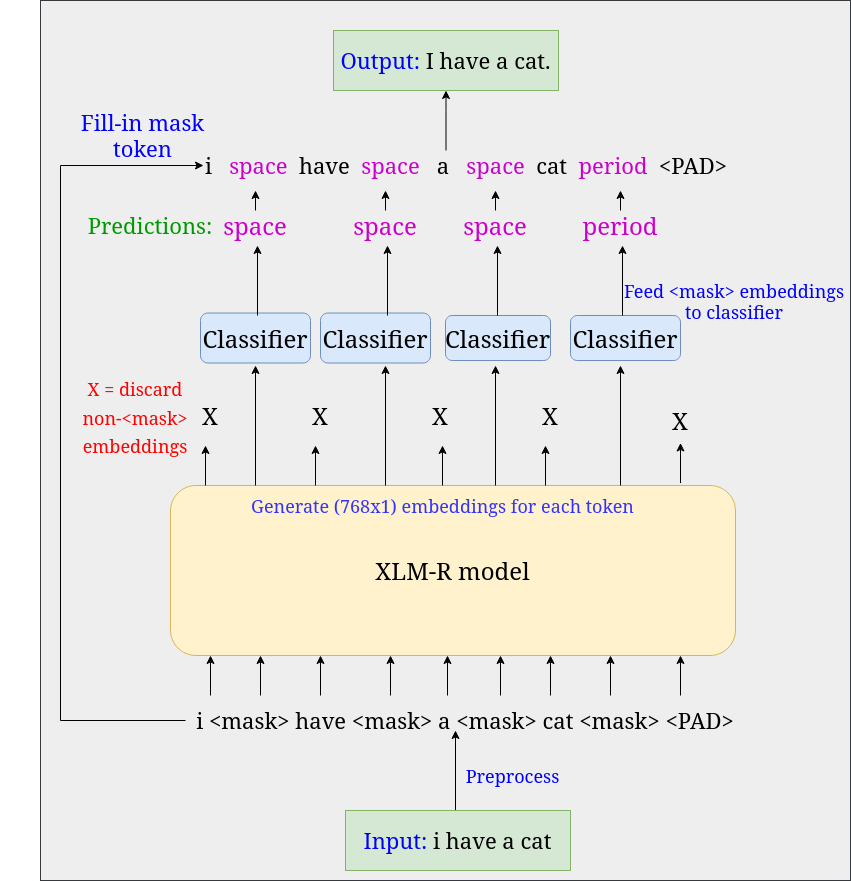} 
	\caption{Schematic diagram of our proposed model architecture}
	\label{fig:model_arch}
\end{figure}

In this work, we propose an architecture similar to that of the MLM pre-training objective of BERT. We use a pre-trained XLM-R model, producing multilingual embeddings, followed by a classifier with two linear layers that output the punctuation for each masked token. 
The proposed architecture is depicted in Figure \ref{fig:model_arch} and more details of the components are described in Section \ref{xlm-r} and Section \ref{classifier}.

\subsection{XLM-R embedding generator}
\label{xlm-r}
The XLM-R model is a variation of the XLM model \cite{DBLP:journals/corr/abs-1901-07291}, which itself enabled better multilingual training from BERT. XLM-R works in a self-supervised fashion and is trained using the Masked-Language Model objective, similar to RoBERTa \cite{DBLP:journals/corr/abs-1907-11692}. XLM-R is trained on 2.5 Terabytes of common-crawled text, while focusing on low-resource languages. Similar to BERT, it contains multiple self-attention transformer-encoder layers stacked on one on top of the other. However, unlike BERT, XLM-R is not trained with the Next Sentence Prediction (NSP) task. Each of these encoder layers consists of transformer multi-head self-attention and feed-forward networks. The self-attention layer, also occasionally referred to as Query-Key-Value attention, maps queries and keys to an output using a scaled dot-product.

\subsection{Classifier}
\label{classifier}
The classifier consists of two Fully-connected (FC) layers. The layers are shared between every mask token for each input sample. The outputs of the classifier of only mask-tokens are considered for back-propagation while others' are discarded, since we only predict punctuation, and the masked tokens represent punctuation marks between two words.
An example of the input and output is described below:\\\\
Input string: \\
\texttt{hello <mask> this <mask> is <mask> josh <mask> here <mask> to <mask> help <mask> you <mask>}\\\\
Output string:\\
\texttt{hello P this O is O josh C here O to O help O you P}\\

The input data has no punctuation and a \texttt{<mask>} token is appended after every word. The output contains $[C, P, Q]$ tokens representing \textit{Comma}, \textit{Period} (full-stop), \textit{Question} for the three punctuation marks that we are predicting and $O$ represents empty space or no-punctuation. 

\subsection{Tokenization}
In order to feed the text into the model, we used the tokenizer called `SentencePiece' provided together with the XLM-R framework to tokenize and encode it into XLM-R input format. The SentencePiece tokenizer is capable of tokenizing multiple languages at once, including English, Malay, and Mandarin.
Punctuation marks that do not conform to the four classes are either modified to the most suitable class or discarded. The English and Malay text are entirely lower-cased to mimic ASR transcripts which usually come without any punctuation or capitalization. 

\begin{figure}
	\centering
	\includegraphics[width=\columnwidth]{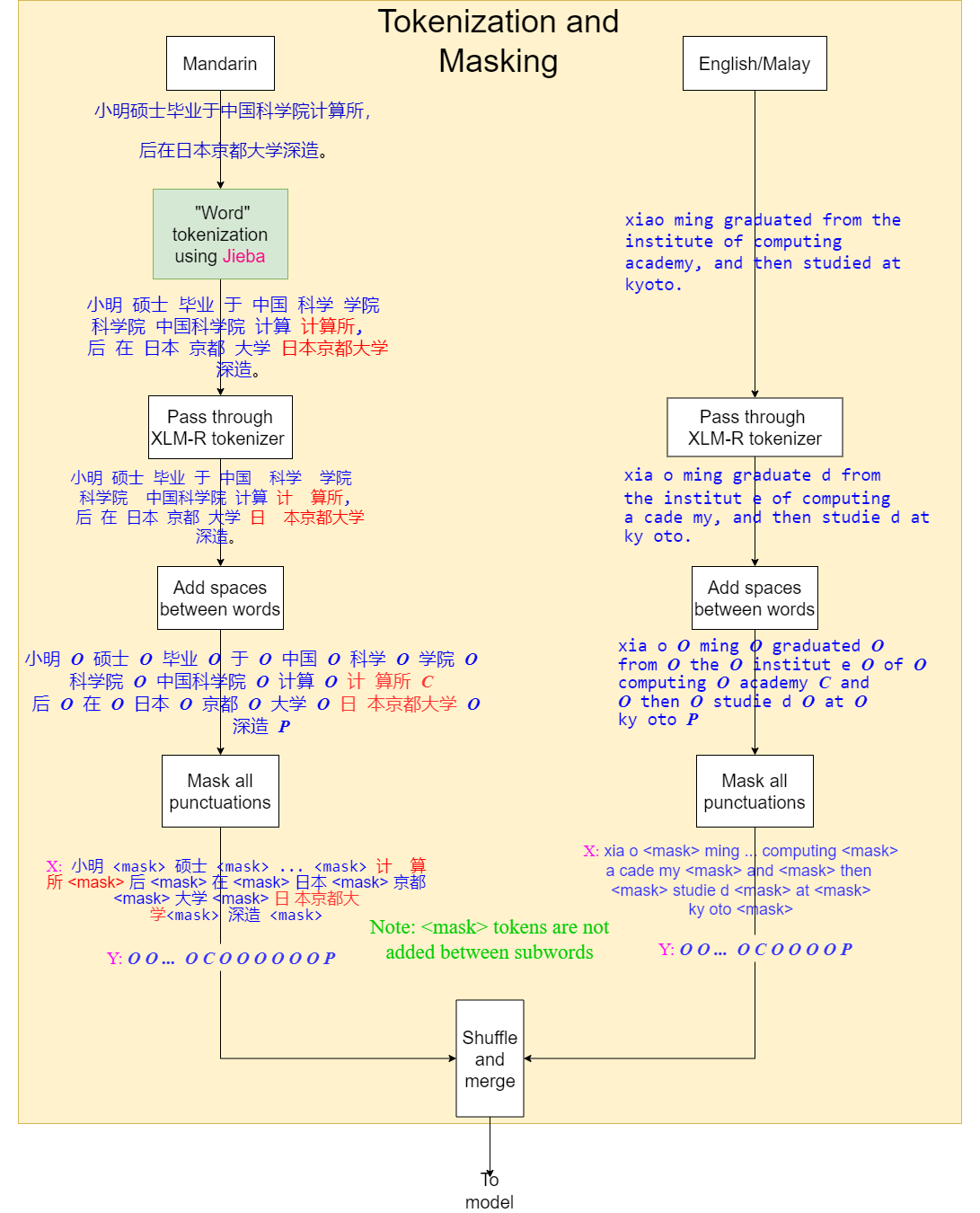}
	\caption{Tokenization and masking process for input data to the model training and evaluation. Note that $[O, C, P]$ are labels for ``Space", ``Comma", and ``Period"}
	\label{fig:preprocessing_train}
\end{figure}

The entire tokenization and masking process for input data is demonstrated in Figure \ref{fig:preprocessing_train}. Specifically, word boundaries are firstly determined, spaces are added between determined words, and subsequently, space or punctuation is replaced with a \texttt{<mask>} token. Word boundaries are easily determined by the space in the case of English and Malay data. For Mandarin, since there is no space between characters or words, we employed $Jieba$\cite{Jieba} as an additional external tokenizer to determine boundaries between each ``word". A space token $O$ or a punctuation token $[C,Q,P]$ is then added between the determined ``words" for the input sequences. 
During training, a \texttt{<mask>} token replaces each space and punctuation token to produce the input sequence $X$, while the accompanying slot-filling label sequence $Y$ is represented by a sequence of $O,C,Q,P$ tokens. 
The entire processed training data, which consists of ${X,Y}$ pairs in English, Malay, and Mandarin, is then shuffled and merged, and subsequently used for model training.
During validation and evaluation, only the $X$ sequences are provided to the model, and there is no shuffling and merging after masking.


\section{Data and Data Pre-procesisng}
\subsection{Datasets}
We train and evaluate our model on monolingual datasets from English, Mandarin, and Malay. For English and Mandarin, we use the IWSLT2012 \cite{DBLP:conf/iwslt/2012} datasets. They consist of transcribed TED talks for Automatic Speech Recognition (ASR), Speech-Language Translation (SLT), and Machine Translation (MT). We consider the MT track for our use case, choosing the EN-FR parallel tracks for English, and the ZH-EN parallel tracks for Mandarin. We use the same train, validation, and test sets from the IWSLT 2012 dataset, similar to previous work on punctuation restoration such as \cite{bertpunct}, \cite{tilk2016bidirectional}, \cite{SAPR} for English and \cite{ZheZhang} for Mandarin. 

Since there is no IWSLT track for Malay, or any publicly available well-formed Malay-transcript dataset, we crawled news articles that were made public by \textit{Zolkepli} \cite{Malay-Dataset}, similar to the work on Bangla punctuation by \cite{TransformerLow}. The dataset is well-formed in sentences and punctuation, and consists of 30,000 news articles crawled on Google News using keyword-matching to find news articles written in Malay, for instance, ``saham malaysia", ``ekonomi malaysia", ``hiburan". We then normalized the crawled Malay text, the details are presented in Section \ref{sec:text_norm}.

\subsection{Data Preprocessing}
\subsubsection{Text Normalization}
\label{sec:text_norm}
The selected English and Mandarin datasets have primarily well-formed texts that require little to no normalization. Similar to \cite{DBLP:journals/corr/ANagy}, we converted or removed punctuation marks that do not belong to the four classes. For instance, we replaced `?!' with `?', and `!' with `.' to mark an end of a sentence. 

The Malay dataset however required additional text normalization to mimic the form of ASR output. Specifically, we have performed a number-to-word conversion, URL address normalization, currency normalization, and phone number conversion. This normalization has been done using the tool developed by \textit{Zolkepli} \cite{Malay-Dataset}.

\subsubsection{Data Sampling for Malay}

The original Malay dataset comes with a huge amount of text with a total 30,000 articles. We have chosen not to use the entire dataset but only retain about 11,000 articles that have more than two sentences in their content. We also noticed that the ratio of question marks is significantly lesser than the other punctuation marks, thus, we have performed oversampling on the number of question marks using random sampling. Specifically, we selected question sentences from the dataset and randomly insert them into parts of the selected articles. Notice that the oversampling has only been performed on the train set, and not on the validation and test sets. After pre-processing, each dataset then has been split into train, validation, and test sets. For English and Mandarin, we adopted the same train, validation, and test sets released by IWSLT. For Malay, we split the dataset with a ratio of 80/10/10 for train, validation, and test sets respectively.

\subsubsection{Final Datasets}

Table \ref{tab:datasets} shows the number of words, commas, question marks, and periods presented in each of the selected datasets after pre-processing. 

\begin{table*}[]
\centering
\caption{Number of words, commas, question marks, and periods presented in train, valid, and test sets}
\resizebox{\textwidth}{!}{%
\begin{tabular}{|l|llll|llll|llll|}
\hline
\multirow{2}{*}{} & \multicolumn{4}{l|}{Train}                                                                              & \multicolumn{4}{l|}{Valid}                                                                      & \multicolumn{4}{l|}{Test}                                                                           \\ \cline{2-13} 
                  & \multicolumn{1}{l|}{Word}      & \multicolumn{1}{l|}{Period}  & \multicolumn{1}{l|}{Question}   & Comma & \multicolumn{1}{l|}{Word} & \multicolumn{1}{l|}{Period} & \multicolumn{1}{l|}{Question} & Comma & \multicolumn{1}{l|}{Word}    & \multicolumn{1}{l|}{Period} & \multicolumn{1}{l|}{Question}  & Comma \\ \hline
English           & \multicolumn{1}{l|}{2,374,992} & \multicolumn{1}{l|}{190,510} & \multicolumn{1}{l|}{10,332} & 141,628    & \multicolumn{1}{l|}{15141}     & \multicolumn{1}{l|}{909}       & \multicolumn{1}{l|}{71}      &     1274     & \multicolumn{1}{l|}{16,208}  & \multicolumn{1}{l|}{1100}   & \multicolumn{1}{l|}{46}     & 1185     \\ \hline
Mandarin           & \multicolumn{1}{l|}{2,321,146} & \multicolumn{1}{l|}{156,373} & \multicolumn{1}{l|}{10,957}  & 85,881   & \multicolumn{1}{l|}{15099}     & \multicolumn{1}{l|}{446}       & \multicolumn{1}{l|}{89}      &    1227      & \multicolumn{1}{l|}{15,099}  & \multicolumn{1}{l|}{446}    & \multicolumn{1}{l|}{89}   & 1227     \\ \hline
Malay             & \multicolumn{1}{l|}{3,135,762} & \multicolumn{1}{l|}{266,108} & \multicolumn{1}{l|}{9,198}   & 252,521  & \multicolumn{1}{l|}{428,959}     & \multicolumn{1}{l|}{24,462}       & \multicolumn{1}{l|}{837}      &  26,023   & \multicolumn{1}{l|}{449,694} & \multicolumn{1}{l|}{26,440} & \multicolumn{1}{l|}{979} &   27,045    \\ \hline
\end{tabular}%
}
\label{tab:datasets}
\end{table*}

\section{Experiments}
\label{experiment}
\subsection{Experimental Setup}
We set up our experiment with the following settings:
\begin{itemize}
\item The classifier comprises two linear layers with dimensions of 1568 and 4, respectively. 
\item Learning rates were set at 1e-4 for the classification layers and 3e-5 for the XLM-R model which is a similar setting to work presented in \cite{DBLP:journals/corr/ANagy}. 
\item We use the RAdam \cite{Liu2020On} optimizer, with a warm-up of 300. 
\item The training data contains English, Mandarin, and Malay datasets and were shuffled before training.
\end{itemize}
The final model was obtained after approximately 9000 iterations, with a batch size of 4 and a sequence length of 256 on a cluster with an Nvidia RTX-3070 and a Tesla-V100 GPUs.

\subsection{Experiment Results}
We report the performance of our model with Precision, Recall, and F1-score for the Period, Comma, and Question mark classes. We compared our model performance against baselines that are three recent works on the task, i.e. SAPR\cite{SAPR}, BertPunct\cite{bertpunct}, and BERT-CNN-RNN\cite{ZheZhang}. The models are briefly described below:
\begin{itemize}
	\item \textbf{SAPR}: This architecture by \textit{Wang et al.} \cite{SAPR} treats the problem as a machine translation task, using a transformer decoder as the base architecture. The work only focuses on English.
	
	\item \textbf{BertPunct}: This work by \textit{Makhija et al.} \cite{bertpunct} treats the problem as a sequence labeling task, using a BERT layer as an embedding and a BiLSTM+CRF layer for classification. This work also focuses only on English. We use the BertPunct\textsubscript{BASE} model for comparison, since we use a similar base model in our work. 
	\item \textbf{BERT-CNN-RNN}: This model proposed by \textit{Zhang et al.} \cite{ZheZhang} focuses on Mandarin. The architecture involves the use of two layers, a word-CNN layer to capture word boundary information, and a layer compromising of BERT-Chinese and an RNN layer for classification of the input into the punctuation class.
    \item \textbf{MLM-Punct}: Our multilingual model. The model was obtained by performing multilingual training with datasets being shuffled for all languages. 
\end{itemize}

\subsubsection{Performance on English test set}

\begin{table}[ht]
\scriptsize
\caption{Performance on English IWSLT2012 Test Set}
\resizebox{\columnwidth}{!}{
\begin{tabular}{|l|l|l|l|l|}
\hline
\multirow{2}{*}{Model} & Period         & Comma           & Question       & Overall                \\ \cline{2-5} 
                       & PR/RC/F1       & PR/RC/F1        & PR/RC/F1       & PR/RC/F1                     \\ \hline
SAPR                   & \textbf{96.7/97.3/96.8} & 57.2/50.8/ 55.9 & 70.6/69.2/70.3 & 78.2/74.4/77.4\\ \hline
BertPunct               & 82.6/83.5/83.1 & \textbf{72.1}/72.4/\textbf{72.3}  & 77.4/\textbf{89.1/82.8} & 77.4/81.7/\textbf{79.4} \\ \hline
MLM-Punct              & 76.8/\textbf{86.7}/81.5 & 60.4/\textbf{78.5}/68.2  & \textbf{78.4}/81.7/80.0 & 68.5/82.4/74.7    \\ \hline
\end{tabular}}
\label{table:en_perform}
\end{table}

Table \ref{table:en_perform} shows performance of SAPR\cite{SAPR}, BertPunct\cite{bertpunct} and MLM-Punct on IWSLT2012 English test set. Note that SAPR and BertPunct are monolingual models while MLM-Punct is multilingual. MLM-Punct's performance is close to the BertPunct model on all three classes with the highest Recall score for Comma and Precision score for the Question mark, while the BertPunct performs better on the rest. SAPR performs well on predicting period which is significantly higher on Precision, Recall, and F1-score for this class, however, it shows poorer performance on \textit{Comma} and \textit{Question} classes.
The difference in performance is possibly due to the English-only BERT encoder providing better word embeddings for English tokens, as opposed to the multilingual XLM-R encoder \cite{pires2019multilingual}. English is also a very high-resource language compared to other languages, and the much larger amounts of English data that go into pre-training English BERT could be another factor.

\subsubsection{Performance on Mandarin test set}
Table \ref{table:cn_perform} shows the performance of our model, MLM-Punct, and the baseline model, BERT-CNN-RNN, on the IWSLT2012 Mandarin test set. Compared to BERT-CNN-RNN, MLM-Punct performs better overall, i.e. by 4.7\% in F1-score. It specifically outperforms on Comma and Period classes with an approximate 17\% and 5\% improvement in F1-score respectively.
\begin{table}[ht]
\scriptsize
\caption{Performance on Mandarin IWSLT2012 Test Set}
\resizebox{\columnwidth}{!}{
\begin{tabular}{|l|l|l|l|l|}
\hline
\multirow{2}{*}{Model} & Period         & Comma          & Question       & Overall \\ \cline{2-5} 
                       & PR/RC/F1       & PR/RC/F1       & PR/RC/F1       & PR/RC/F1 \\ \hline
Zhang et al.           & 68.4/\textbf{83.5}/75.2 & 52.0/60.1/55.8 & \textbf{78.8}/73.9/76.3 & 66.4/72.5/69.1\\ \hline
MLM-Punct              & \textbf{73.8}/80.3/\textbf{76.9} & \textbf{66.3/80.0/72.5} & 75.0/\textbf{88.7/81.3} & \textbf{68.6/80.5/73.8} \\ \hline
\end{tabular}}
\label{table:cn_perform}
\end{table}
\subsubsection{Performance on Malay test set}
To the best of our knowledge, there is no publicly available punctuation restoration model for Malay language. Our model performs well for $Period$ and $Question$  classes, and slightly lower F1-score for $Comma$ class with F1-scores at 86.1\%, 80.0\% and 70.8\% for each class respectively.
\begin{table}[ht]
\scriptsize
\caption{Performance on Malay News Test Set}
\resizebox{\columnwidth}{!}{
\begin{tabular}{|l|l|l|l|l|}
\hline
\multirow{2}{*}{Model} & Period         & Comma            & Question       & Overall\\ \cline{2-5} 
                       & PR/RC/F1       & PR/RC/F1         & PR/RC/F1       & PR/RC/F1\\ \hline
MLM-Punct              & 81.0/91.9/86.1 &  61.7/83.0/70.8 & 73.6/87.6/80.0 & 70.8/87.2/78.0\\ \hline
\end{tabular}}
\label{table:3}
\end{table}
\section{Discussion}
\label{discussion}
In addition to building a multilingual model that can restore punctuation for the three languages at once, we have experimented with building monolingual models for each language. We present a comparison between these models in Section \ref{comparison}. Subsequently, Section \ref{jieba} discusses the effectiveness of using the Jieba tokenizer on Mandarin text.
\subsection{A Comparison between Monolingual and Multilingual training}
\label{comparison}
Table \ref{tab:4} presents the experimental results on monolingual and multilingual training settings. Monolingual models, i.e. Mono-English, Mono-Mandarin, Mono-Malay were trained on the same network architecture as MLM-Punct with pre-trained multilingual XLM-RoBERTa. However, they were fine-tuned by using only the monolingual dataset of the respective language i.e. English IWSLT2012, Mandarin IWSLT2012, and Malay News.
The experimental results show that MLM-Punct outperforms monolingual training across three test sets on overall Precision, Recall, and F1 scores. Specifically, MLM-Punct gains 1.6\%, 1.6\% and 3.6\% of F1 scores on English, Mandarin, and Malay test sets respectively.

The gains in the overall performance of the multilingual model are possibly attributed to XLM-R's cross-lingual pre-training that brings out linguistic similarities across languages when the model is fine-tuned on a multilingual dataset. Another possibility is the cross-lingual nature of XLM-R shares the latent properties relating to punctuation and phrasal information across languages and thus benefit each other.
However, there are some dips in some specific classes. For example, The Precision score is dipped for $Period$ class on Mandarin and Malay test sets. Another dip is from the $Question$ class in Mandarin, which shows an F1-score dip of 1.6\% as compared to the Mono-Mandarin model.

\begin{table*}[!htbp]{
\centering
\caption{Results on monolingual and multilingual training. MLM-Punct represents multilingual training. Mono-X represents model fine-tuned on Language `X'}
\resizebox{\textwidth}{!}{%
\begin{tabular}{|l|llll|llll|llll|}
\hline
\multirow{3}{*}{Models} & \multicolumn{4}{l|}{English IWSLT Test set}                                                                                                                                                  & \multicolumn{4}{l|}{Mandarin IWSLT Test set}                                                                                                                                                           & \multicolumn{4}{l|}{Malay news Test set}                                                                                                                                                    \\ \cline{2-13} 
                           & \multicolumn{1}{l|}{Period}                  & \multicolumn{1}{l|}{Comma}                            & \multicolumn{1}{l|}{Question}                & Overall                 & \multicolumn{1}{l|}{Period}                  & \multicolumn{1}{l|}{Comma}                            & \multicolumn{1}{l|}{Question}                         & Overall                 & \multicolumn{1}{l|}{Period}                  & \multicolumn{1}{l|}{Comma}                            & \multicolumn{1}{l|}{Question}                & Overall                 \\ 
\cline{2-13}                     & \multicolumn{1}{l|}{PR/RC/F1}                & \multicolumn{1}{l|}{PR/RC/F1}                         & \multicolumn{1}{l|}{PR/RC/F1}                & PR/RC/F1                & \multicolumn{1}{l|}{PR/RC/F1}                & \multicolumn{1}{l|}{PR/RC/F1}                         & \multicolumn{1}{l|}{PR/RC/F1}                         & PR/RC/F1                & \multicolumn{1}{l|}{PR/RC/F1}                & \multicolumn{1}{l|}{PR/RC/F1}                         & \multicolumn{1}{l|}{PR/RC/F1}                & PR/RC/F1                \\ \hline
MLM-Punct               & \multicolumn{1}{l|}{\textbf{76.8/86.7/81.5}} & \multicolumn{1}{l|}{\textbf{60.4}/78.5/\textbf{68.2}} & \multicolumn{1}{l|}{\textbf{78.4/81.7/80.0}} & \textbf{68.5/82.4/74.7} & \multicolumn{1}{l|}{73.8/\textbf{80.3/76.9}} & \multicolumn{1}{l|}{\textbf{66.3}/80.0/\textbf{72.5}} & \multicolumn{1}{l|}{75.0/\textbf{88.7}/76.3}          & \textbf{68.6/80.5/73.8} & \multicolumn{1}{l|}{81.0/\textbf{91.9/86.1}} & \multicolumn{1}{l|}{\textbf{61.7}/83.0/\textbf{70.8}} & \multicolumn{1}{l|}{\textbf{73.6/87.6/80.0}} & \textbf{70.8/87.2/78.0} \\ \hline
Mono-English                    & \multicolumn{1}{l|}{73.6/86.1/79.3}          & \multicolumn{1}{l|}{58.7/\textbf{78.7}/67.3}          & \multicolumn{1}{l|}{74.6/74.6/74.6}          & 66.0/82.1/73.1          & \multicolumn{1}{l|}{-}                       & \multicolumn{1}{l|}{-}                                & \multicolumn{1}{l|}{-}                                & -                       & \multicolumn{1}{l|}{-}                       & \multicolumn{1}{l|}{-}                                & \multicolumn{1}{l|}{-}                       & -                       \\ \hline
Mono-Mandarin                    & \multicolumn{1}{l|}{-}                       & \multicolumn{1}{l|}{-}                                & \multicolumn{1}{l|}{-}                       & -                       & \multicolumn{1}{l|}{\textbf{75.1}/77.6/76.3} & \multicolumn{1}{l|}{60.3/\textbf{82.9}/69.8}          & \multicolumn{1}{l|}{\textbf{81.9}/86.5/\textbf{84.2}} & 65.1/81.7/72.2          & \multicolumn{1}{l|}{-}                       & \multicolumn{1}{l|}{-}                                & \multicolumn{1}{l|}{-}                       & -                       \\ \hline
Mono-Malay                      & \multicolumn{1}{l|}{-}                       & \multicolumn{1}{l|}{-}                                & \multicolumn{1}{l|}{-}                       & -                       & \multicolumn{1}{l|}{-}                       & \multicolumn{1}{l|}{-}                                & \multicolumn{1}{l|}{-}                                & -                       & \multicolumn{1}{l|}{\textbf{83.5}/81.5/82.5} & \multicolumn{1}{l|}{55.8/\textbf{85.0}/67.4}          & \multicolumn{1}{l|}{69.2/58.3/63.2}          & 68.9/83.1/74.4          \\ \hline
\end{tabular}%
}
\label{tab:4}}
\end{table*}

\subsection{Using Jieba tokenization for Mandarin}
\label{jieba}
Mandarin text usually comes with no space between characters. Thus, we first have to split the text into ``word" tokens and then insert \textit{[SPACE]} tokens after every token. The \textit{[SPACE]} tokens and the punctuation marks are masked before feeding the input to the model.
To split the text, we have experimented with using only the XLM-R built-in SentencePiece tokenizer and by using $Jieba$ tokenizer \cite{Jieba}. Our experimental results show that using $Jieba$ helps to improve the model performance significantly.

The experiment results are presented in Table \ref{table:5}. We named the model using the SentencePiece tokenizer as $MLM-Punct_s$, while our $MLM-Punct$ model represents the model using $Jieba$ tokenizer. $MLM-Punct$ outperforms $MLM-Punct_s$ on the Mandarin test set with 17.2\% improvement on F1-score. This is probably because $Jieba$ tokenizes Mandarin text into ``words", while SentencePiece tends to tokenize into sub-words. Predicting punctuation marks between words is more syntactically and semantically accurate than predicting punctuation marks between sub-words.
In \cite{rust2021good}, the authors reported a similar issue in dealing with Mandarin text in using WordPiece tokenizer for BERT. Specifically, it tends to perform poorly for languages with non-trivial word boundaries on the tasks that involve word-level information such as POS tagging and NER.
\begin{table}[!htbp]
\caption{Comparison between using SentencePiece tokenizer and $Jieba$ tokenizer for Mandarin data. $MLM-Punct_s$ represents the model using SentencePiece tokenizer and MLM-Punct represents the model using $Jieba$ tokenizer}
\resizebox{\columnwidth}{!}{
\begin{tabular}{|l|l|l|l|l|}
\cline{1-5}
\multirow{2}{*}{Model} & Period         & Comma          & Question       &  Overall\\ \cline{2-5}
                             & PR/RC/F1       & PR/RC/F1       & PR/RC/F1       &  PR/RC/F1\\ \cline{1-5}
$MLM-Punct_{s}$          & 34.2/84.4/48.7 & 47.1/75.1/57.9 & 70.3/87.4/77.9 &  45.0/78.1/56.6\\ \cline{1-5}
$MLM-Punct$              & 73.8/80.3/76.9 & 66.3/80.0/72.5 & 75.0/88.7/76.3 &  \textbf{68.6/80.5/73.8}\\ \cline{1-5}
\end{tabular}}
\label{table:5}
\end{table}

\section{Conclusions}
\label{conclusion}
In this work, our proposed model has the ability to punctuate multilingual text in English, Mandarin, and Malay, treating the problem as a slot-filling task using the XLM-RoBERTa language representation model. We show that multilingual training can perform better than monolingual training and achieve state-of-the-art results for Mandarin punctuation.

For future work, we will consider acquiring more standard evaluation sets for English and Mandarin, considering that the IWSLT track mainly involves describing TED Talks. Malay does not have a proper speech transcription dataset, and requires additional dataset procurement. We will also train and evaluate on a larger set of languages. 
Another direction would be to incorporate prosody features into the architecture to predict uncommon punctuation. A further contribution following this could be on-the-fly punctuation in the automatic speech recognition pipeline for the multilingual space. Previous work exists for the English language \cite{chen1999speech}, \cite{yuanPunctASR}, \cite{hlubik2020inserting}, but there has been currently been no work towards the multilingual space.
Finally, we could explore the task of punctuating code-switching text, which is a challenge due to no defined grammatical rules surrounding them, thus being a purely machine-learning-oriented task. 

\section{Acknowledgments}
This research is supported by ST Engineering Mission Software \& Services Pte. Ltd under a collaboration programme (Research Collaboration No: REQ0149132). The computational work for this paper was partially performed on resources of the National Supercomputing Center, Singapore (https://www.nscc.sg), and on Sharanga, BITS Pilani, Hyderabad Campus.\\

\bibliographystyle{IEEEtran}

\bibliography{mybib}

\begin{thebibliography}{10}
\providecommand{\url}[1]{#1}
\csname url@samestyle\endcsname
\providecommand{\newblock}{\relax}
\providecommand{\bibinfo}[2]{#2}
\providecommand{\BIBentrySTDinterwordspacing}{\spaceskip=0pt\relax}
\providecommand{\BIBentryALTinterwordstretchfactor}{4}
\providecommand{\BIBentryALTinterwordspacing}{\spaceskip=\fontdimen2\font plus
\BIBentryALTinterwordstretchfactor\fontdimen3\font minus
  \fontdimen4\font\relax}
\providecommand{\BIBforeignlanguage}[2]{{%
\expandafter\ifx\csname l@#1\endcsname\relax
\typeout{** WARNING: IEEEtran.bst: No hyphenation pattern has been}%
\typeout{** loaded for the language `#1'. Using the pattern for}%
\typeout{** the default language instead.}%
\else
\language=\csname l@#1\endcsname
\fi
#2}}
\providecommand{\BIBdecl}{\relax}
\BIBdecl

\bibitem{jones-et-al-punctutation}
D.~Jones, F.~Wolf, E.~Gibson, E.~Williams, E.~Fedorenko, D.~Reynolds, and
  M.~Zissman, ``Measuring the readability of automatic speech-to-text
  transcripts,'' 01 2003.

\bibitem{IWSLT2011/MT}
S.~Peitz, M.~Freitag, A.~Mauser, and H.~Ney, ``Modeling punctuation prediction
  as machine translation,'' 12 2011, pp. 238--245.

\bibitem{spitkovsky-etal-2011-punctuation}
\BIBentryALTinterwordspacing
V.~I. Spitkovsky, H.~Alshawi, and D.~Jurafsky, ``{P}unctuation: Making a point
  in unsupervised dependency parsing,'' in \emph{Proceedings of the Fifteenth
  Conference on Computational Natural Language Learning}.\hskip 1em plus 0.5em
  minus 0.4em\relax Portland, Oregon, USA: Association for Computational
  Linguistics, Jun. 2011, pp. 19--28. [Online]. Available:
  \url{https://aclanthology.org/W11-0303}
\BIBentrySTDinterwordspacing

\bibitem{Annotate}
H.~Christensen, Y.~Gotoh, and S.~Renals, ``Punctuation annotation using
  statistical prosody models,'' \emph{Prosody and Speech Recognition}, 10 2001.

\bibitem{Batista2007RecoveringPM}
F.~Batista, D.~Caseiro, N.~J. Mamede, and I.~Trancoso, ``Recovering punctuation
  marks for automatic speech recognition,'' in \emph{INTERSPEECH}, 2007.

\bibitem{tilk2015lstm}
O.~Tilk and T.~Alum{\"a}e, ``Lstm for punctuation restoration in speech
  transcripts,'' in \emph{Sixteenth annual conference of the international
  speech communication association}, 2015.

\bibitem{9747131}
Y.~Zhu, L.~Wu, S.~Cheng, and M.~Wang, ``Unified multimodal punctuation
  restoration framework for mixed-modality corpus,'' in \emph{ICASSP 2022 -
  2022 IEEE International Conference on Acoustics, Speech and Signal Processing
  (ICASSP)}, 2022, pp. 7272--7276.

\bibitem{article_HMM}
Y.~Liu, E.~Shriberg, A.~Stolcke, D.~Hillard, M.~Ostendorf, and M.~Harper,
  ``Enriching speech recognition with automatic detection of sentence
  boundaries and disfluencies,'' \emph{Audio, Speech, and Language Processing,
  IEEE Transactions on}, vol.~14, pp. 1526 -- 1540, 10 2006.

\bibitem{HMMInproceedings}
W.~Lu and H.~Ng, ``Better punctuation prediction with dynamic conditional
  random fields.'' \emph{EMNLP 2010 - Conference on Empirical Methods in
  Natural Language Processing, Proceedings of the Conference}, pp. 177--186, 01
  2010.

\bibitem{DBLP:conf/lrec/CheWYM16}
\BIBentryALTinterwordspacing
X.~Che, C.~Wang, H.~Yang, and C.~Meinel, ``Punctuation prediction for
  unsegmented transcript based on word vector,'' in \emph{Proceedings of the
  Tenth International Conference on Language Resources and Evaluation {LREC}
  2016, Portoro{\v{z}}, Slovenia, May 23-28, 2016}, N.~Calzolari, K.~Choukri,
  T.~Declerck, S.~Goggi, M.~Grobelnik, B.~Maegaard, J.~Mariani, H.~Mazo,
  A.~Moreno, J.~Odijk, and S.~Piperidis, Eds.\hskip 1em plus 0.5em minus
  0.4em\relax European Language Resources Association {(ELRA)}, 2016. [Online].
  Available:
  \url{http://www.lrec-conf.org/proceedings/lrec2016/summaries/103.html}
\BIBentrySTDinterwordspacing

\bibitem{devlin2019bert}
J.~Devlin, M.-W. Chang, K.~Lee, and K.~Toutanova, ``Bert: Pre-training of deep
  bidirectional transformers for language understanding,'' 2019.

\bibitem{conneau2020unsupervised}
A.~Conneau, K.~Khandelwal, N.~Goyal, V.~Chaudhary, G.~Wenzek, F.~Guzmán,
  E.~Grave, M.~Ott, L.~Zettlemoyer, and V.~Stoyanov, ``Unsupervised
  cross-lingual representation learning at scale,'' 2020.

\bibitem{DBLP:XNLI}
\BIBentryALTinterwordspacing
A.~Conneau, G.~Lample, R.~Rinott, A.~Williams, S.~R. Bowman, H.~Schwenk, and
  V.~Stoyanov, ``{XNLI:} evaluating cross-lingual sentence representations,''
  \emph{CoRR}, vol. abs/1809.05053, 2018. [Online]. Available:
  \url{http://arxiv.org/abs/1809.05053}
\BIBentrySTDinterwordspacing

\bibitem{DBLP:XQuAD}
\BIBentryALTinterwordspacing
M.~Artetxe, S.~Ruder, and D.~Yogatama, ``On the cross-lingual transferability
  of monolingual representations,'' \emph{CoRR}, vol. abs/1910.11856, 2019.
  [Online]. Available: \url{http://arxiv.org/abs/1910.11856}
\BIBentrySTDinterwordspacing

\bibitem{tiedemann2020tatoeba}
J.~Tiedemann, ``The tatoeba translation challenge--realistic data sets for low
  resource and multilingual mt,'' \emph{arXiv preprint arXiv:2010.06354}, 2020.

\bibitem{tilk2016bidirectional}
O.~Tilk and T.~Alum{\"a}e, ``Bidirectional recurrent neural network with
  attention mechanism for punctuation restoration.'' in \emph{Interspeech},
  2016, pp. 3047--3051.

\bibitem{salloum-etal-2017-deep}
\BIBentryALTinterwordspacing
W.~Salloum, G.~Finley, E.~Edwards, M.~Miller, and D.~Suendermann-Oeft, ``Deep
  learning for punctuation restoration in medical reports,'' in
  \emph{{B}io{NLP} 2017}.\hskip 1em plus 0.5em minus 0.4em\relax Vancouver,
  Canada,: Association for Computational Linguistics, Aug. 2017, pp. 159--164.
  [Online]. Available: \url{https://aclanthology.org/W17-2319}
\BIBentrySTDinterwordspacing

\bibitem{DBLP:journals/corr/VaswaniSPUJGKP17}
\BIBentryALTinterwordspacing
A.~Vaswani, N.~Shazeer, N.~Parmar, J.~Uszkoreit, L.~Jones, A.~N. Gomez,
  L.~Kaiser, and I.~Polosukhin, ``Attention is all you need,'' \emph{CoRR},
  vol. abs/1706.03762, 2017. [Online]. Available:
  \url{http://arxiv.org/abs/1706.03762}
\BIBentrySTDinterwordspacing

\bibitem{bertpunct}
K.~Makhija, T.-N. Ho, and E.-S. Chng, ``Transfer learning for punctuation
  prediction,'' in \emph{2019 Asia-Pacific Signal and Information Processing
  Association Annual Summit and Conference (APSIPA ASC)}, 2019, pp. 268--273.

\bibitem{yi2020adversarial}
J.~Yi, J.~Tao, Y.~Bai, Z.~Tian, and C.~Fan, ``Adversarial transfer learning for
  punctuation restoration,'' 2020.

\bibitem{DiscriminQian}
\BIBentryALTinterwordspacing
Q.~Chen, W.~Wang, M.~Chen, and Q.~Zhang, ``Discriminative self-training for
  punctuation prediction,'' \emph{Interspeech 2021}, Aug 2021. [Online].
  Available: \url{http://dx.doi.org/10.21437/Interspeech.2021-246}
\BIBentrySTDinterwordspacing

\bibitem{ZheZhang}
Z.~Zhang, J.~Liu, L.~Chi, and X.~Chen, ``Word-level bert-cnn-rnn model for
  chinese punctuation restoration,'' in \emph{2020 IEEE 6th International
  Conference on Computer and Communications (ICCC)}, 2020, pp. 1629--1633.

\bibitem{SAPR}
F.~Wang, W.~Chen, Z.~Yang, and B.~Xu, ``Self-attention based network for
  punctuation restoration,'' in \emph{2018 24th International Conference on
  Pattern Recognition (ICPR)}, 2018, pp. 2803--2808.

\bibitem{TransformerLow}
T.~Alam, A.~Khan, and F.~Alam, ``Punctuation restoration using transformer
  models for high-and low-resource languages,'' in \emph{Proceedings of the
  Sixth Workshop on Noisy User-generated Text (W-NUT 2020)}, 01 2020, pp.
  132--142.

\bibitem{DBLP:journals/corr/ANagy}
\BIBentryALTinterwordspacing
A.~Nagy, B.~Bial, and J.~{\'{A}}cs, ``Automatic punctuation restoration with
  {BERT} models,'' \emph{CoRR}, vol. abs/2101.07343, 2021. [Online]. Available:
  \url{https://arxiv.org/abs/2101.07343}
\BIBentrySTDinterwordspacing

\bibitem{chordia-2021-punktuator}
\BIBentryALTinterwordspacing
V.~Chordia, ``{P}un{K}tuator: A multilingual punctuation restoration system for
  spoken and written text,'' in \emph{Proceedings of the 16th Conference of the
  European Chapter of the Association for Computational Linguistics: System
  Demonstrations}.\hskip 1em plus 0.5em minus 0.4em\relax Online: Association
  for Computational Linguistics, Apr. 2021, pp. 312--320. [Online]. Available:
  \url{https://aclanthology.org/2021.eacl-demos.37}
\BIBentrySTDinterwordspacing

\bibitem{GUERREIRO2021115740}
\BIBentryALTinterwordspacing
N.~M. Guerreiro, R.~Rei, and F.~Batista, ``Towards better subtitles: A
  multilingual approach for punctuation restoration of speech transcripts,''
  \emph{Expert Systems with Applications}, vol. 186, p. 115740, 2021. [Online].
  Available:
  \url{https://www.sciencedirect.com/science/article/pii/S0957417421011180}
\BIBentrySTDinterwordspacing

\bibitem{li20m_interspeech}
X.~Li and E.~Lin, ``{A 43 Language Multilingual Punctuation Prediction Neural
  Network Model},'' in \emph{Proc. Interspeech 2020}, 2020, pp. 1067--1071.

\bibitem{DBLP:journals/corr/abs-1901-07291}
\BIBentryALTinterwordspacing
G.~Lample and A.~Conneau, ``Cross-lingual language model pretraining,''
  \emph{CoRR}, vol. abs/1901.07291, 2019. [Online]. Available:
  \url{http://arxiv.org/abs/1901.07291}
\BIBentrySTDinterwordspacing

\bibitem{DBLP:journals/corr/abs-1907-11692}
\BIBentryALTinterwordspacing
Y.~Liu, M.~Ott, N.~Goyal, J.~Du, M.~Joshi, D.~Chen, O.~Levy, M.~Lewis,
  L.~Zettlemoyer, and V.~Stoyanov, ``Roberta: {A} robustly optimized {BERT}
  pretraining approach,'' \emph{CoRR}, vol. abs/1907.11692, 2019. [Online].
  Available: \url{http://arxiv.org/abs/1907.11692}
\BIBentrySTDinterwordspacing

\bibitem{Jieba}
S.~Junyi, ``Jieba, word tokenizer for chinese,''
  \url{https://github.com/fxsjy/jieba}, 2013.

\bibitem{DBLP:conf/iwslt/2012}
\BIBentryALTinterwordspacing
\emph{2012 International Workshop on Spoken Language Translation, {IWSLT} 2012,
  Hong Kong, December 6-7, 2012}.\hskip 1em plus 0.5em minus 0.4em\relax
  {ISCA}, 2012. [Online]. Available:
  \url{http://www.isca-speech.org/archive/iwslt\_12/}
\BIBentrySTDinterwordspacing

\bibitem{Malay-Dataset}
Z.~Husein, ``Malay-dataset,''
  \url{https://github.com/huseinzol05/malay-dataset}, 2018.

\bibitem{Liu2020On}
\BIBentryALTinterwordspacing
L.~Liu, H.~Jiang, P.~He, W.~Chen, X.~Liu, J.~Gao, and J.~Han, ``On the variance
  of the adaptive learning rate and beyond,'' in \emph{International Conference
  on Learning Representations}, 2020. [Online]. Available:
  \url{https://openreview.net/forum?id=rkgz2aEKDr}
\BIBentrySTDinterwordspacing

\bibitem{pires2019multilingual}
T.~Pires, E.~Schlinger, and D.~Garrette, ``How multilingual is multilingual
  bert?'' 2019.

\bibitem{rust2021good}
P.~Rust, J.~Pfeiffer, I.~Vulić, S.~Ruder, and I.~Gurevych, ``How good is your
  tokenizer? on the monolingual performance of multilingual language models,''
  2021.

\bibitem{chen1999speech}
C.~J. Chen, ``Speech recognition with automatic punctuation,'' in \emph{Sixth
  European Conference on Speech Communication and Technology}, 1999.

\bibitem{yuanPunctASR}
\BIBentryALTinterwordspacing
Y.~Guan, ``End to end asr system with automatic punctuation insertion,'' 2020.
  [Online]. Available: \url{https://arxiv.org/abs/2012.02012}
\BIBentrySTDinterwordspacing

\bibitem{hlubik2020inserting}
P.~Hlub{\'\i}k, M.~{\v{S}}pan{\v{e}}l, M.~Boh{\'a}{\v{c}}, and
  L.~Weingartov{\'a}, ``Inserting punctuation to asr output in a real-time
  production environment,'' in \emph{International Conference on Text, Speech,
  and Dialogue}.\hskip 1em plus 0.5em minus 0.4em\relax Springer, 2020, pp.
  418--425.

\end{thebibliography}


\begin{thebibliography}{1}

\bibitem{1}
G.~Eason, B.~Noble, and I.~N.~Sneddon, ``On certain integrals of
Lipschitz-Hankel type involving products of Bessel functions,''
\emph{Phil. Trans. Roy. Soc. London,} vol. A247, pp. 529-551, April
1955.

\bibitem{2}
J.~Clerk~Maxwell, \emph{A Treatise on Electricity and Magnetism,}
3$^{\rm rd}$ ed., vol. 2. Oxford: Clarendon, 1892, pp.68-73.

\bibitem{3}
I.~S.~Jacobs and C.~P.~Bean, ``Fine particles, thin films and exchange
anisotropy,'' in \emph{Magnetism,} vol. III, G.T. Rado and H. Suhl,
Eds. New York: Academic, 1963, pp. 271-350.

\bibitem{4}
K.~Elissa, ``Title of paper if known,'' unpublished.

\bibitem{5}
R.~Nicole, ``Title of paper with only first word capitalized,''
\emph{J. Name Stand. Abbrev.,} in press.

\bibitem{6}
Y.~Yorozu, M.~Hirano, K.~Oka, and Y.~Tagawa, ``Electron spectroscopy
studies on magneto-optical media and plastic substrate interface,''
\emph{APSIPA Transl. J. Magn. Japan,} vol. 2, pp. 740-741, August 1987
[\emph{Digests 9$^{\rm th}$ Annual Conf. Magnetics Japan,} p. 301,
1982].

\bibitem{7}
M.~Young, \emph{The Technical Writer's Handbook.} Mill Valley, CA:
University Science, 1989.

\end{thebibliography}

\end{document}